\documentclass{article} % For LaTeX2e
\usepackage{iclr2022_conference,times}
\usepackage{graphicx} % DO NOT CHANGE THIS

\usepackage{amsmath,amsfonts,bm}

\def\eqref#1{equation~\ref{#1}}
\def\1{\bm{1}}

\DeclareMathAlphabet{\mathsfit}{\encodingdefault}{\sfdefault}{m}{sl}
\SetMathAlphabet{\mathsfit}{bold}{\encodingdefault}{\sfdefault}{bx}{n}

\usepackage{hyperref}
\usepackage{url}
\usepackage{wrapfig}
\usepackage{multirow}
\usepackage{enumitem}

\usepackage{subcaption}

\title{BERMo:  What can \textbf{BER}T learn from EL\textbf{Mo}?}

\author{Sangamesh Kodge \& Kaushik Roy \\
School of Electrical and Computer Engineering, Purdue University\\
\texttt{\{skodge,kuashik\}@purdue.edu} \\
}

\iclrfinalcopy % Uncomment for camera-ready version, but NOT for submission.
\begin{document}

\maketitle

\begin{abstract}
    We propose BERMo, an architectural modification to BERT, which makes predictions based on a hierarchy  of surface, syntactic and semantic language features. 
    We use linear combination scheme proposed in Embeddings from Language Models (ELMo) to combine the scaled internal representations from different network depths. 
    Our approach has two-fold benefits: (1) improved gradient flow for the downstream task as every layer has a direct connection to the gradients of the loss function and (2) increased representative power as the model no longer needs to copy the features learned in the shallower layer which are necessary for the downstream task. 
    Further,  our model has a negligible parameter overhead as there is a single scalar parameter associated with each layer in the network. 
    Experiments on the probing task from SentEval dataset show that our model performs up to $4.65\%$ better in accuracy than the baseline with an average improvement of $2.67\%$ on the semantic tasks.
    When subject to compression techniques, we find that our model enables stable pruning for compressing small datasets like SST-2, where the BERT model commonly diverges. 
    We observe that our approach converges $1.67\times$ and $1.15\times$ faster than the baseline on MNLI and QQP tasks from GLUE dataset. 
    Moreover, our results show that our approach can obtain better parameter efficiency for penalty based pruning approaches on QQP task.
\end{abstract}

    % Prevoious version
    % This paper introduces BERMo, an architectural modification to BERT, which makes predictions based on a hierarchy  of surface, syntactic and semantic language features. We use the linear combination scheme proposed in Embeddings from Language Models (ELMo) to combine the scaled internal representations from different network depths. As there is a single scalar parameter associated with each layer in the network, our model has a negligible parameter overhead. Further, our approach has two-fold benefits: (1) improved gradient flow for the downstream task as every layer has a direct connection to the gradients of the loss function and (2) increased representative power as the model no longer needs to copy the features learned in the shallower layer which are necessary for the downstream task. Our experiments on the probing task from the SentEval dataset show that our model performs up to $4.65\%$ better in accuracy than the baseline with an average improvement of $2.67\%$ on the semantic tasks. When subject to compression techniques, we find that our model enables stable pruning for compressing small datasets like SST-2, where the BERT-BASE model commonly diverges. Moreover,  we observe that our approach converges $1.67\times$ and $1.15\times$ faster than the baseline on MNLI and QQP tasks from the GLUE dataset.
%\newpage

\section{Introduction}
%\noindent \textcolor{blue}{Why looking at bert is important?} %%% Probably not important!
\par \noindent 
    The invention of Transformer (\cite{transformer}) architecture has paved new research directions in the deep learning community. 
    Descendants of this architecture, namely BERT (\cite{bert}) and  GPT (\cite{gpt}), attain State-of-The-Art (SoTA) performance for a broad range of NLP applications. 
    The success of these networks is primarily attributed to the two-stage training process (self-supervised pre-training and task-based fine-tuning), 
    %first introduced in Embeddings from Language Models (ELMo) (\cite{elmo}) 
    and the attention mechanism introduced in Transformers. 
    Many of the top models on various leader boards use models from the BERT family. 
    All of the fifteen systems that surpass the human baseline on the General Language Understanding Evaluation (GLUE) (\cite{glue}) benchmark use variants of BERT or have it as one of the constituents in an ensemble, except for T5 (\cite{t5}), which uses the Transformer architecture. 
    Further, the best-performing systems for each task in the Ontonotes (\cite{ontonotes}) benchmark belong to the BERT family, with the exception of Entity Typing task where Embeddings from Language Models (ELMo) (\cite{elmo}) tops the leaderboard. 
    These promising results make the BERT family of models increasingly ubiquitous in solving several tasks in the various domains of machine learning like NLP, Image Recognition (\cite{dosovitskiy2021image, jaegle2021perceiver}) and Object detection (\cite{10.1007/978-3-030-58452-8_13}).

%\noindent \textcolor{blue}{Motivation/intuition for adding skip connections}
\par 
    \textbf{Motivation: }In (\cite{bert-language-structure}), the authors manifest the lower layers of BERT to capture phrase-level information, which gets diluted with the depth. Moreover, they also demonstrate that the initial layers capture surface-level features, the middle layers deal with syntactic features, and the last few layers are responsible for semantic features. These findings indicate that BERT captures a rich hierarchy of linguistic features at different depths of the network. Intrigued by this discovery, we aim at combining the activations from different depths to obtain a richer feature representation. We find our problem formulation similar to the one presented in ELMo, where the authors illustrate higher-level Long Short Term Memory (LSTM) states capture context-dependent aspects of word meaning or semantic features, and the lower-level LSTM states model the syntax. 
    Inspired by ELMo, we propose BERMo by modifying the BERT architecture to increase the dependence of features from different depths to generate a rich context-dependent embedding. This approach improves the gradient flow during the backward pass and increases the representative power of the network (\cite{resnet, DBLP:journals/corr/HuangLW16a}).   
%\noindent \textcolor{blue}{Motivation for Compression}
%\par 
    Further, the linear combination of features from intermediate layers, proposed in ELMo, is simpler form of skip connection introduced in ResNets (\cite{resnet}). The skip connections in ResNets enable aggressive pooling in initial layers without affecting the gradient flow. This in turn leads to low parameters allowing these networks to have orders of magnitude lower parameters compared to the architectures without skip connections such as VGG (\cite{vgg}) while achieving competitive accuracy. Since the performance improvements associated with the BERT architecture come at the cost of a large memory footprint and enormous computational resources, compression becomes necessary to deploy these networks in resource-constrained environments like drones, mobile computers and IoT devices. As we introduce skip connections in the architecture to obtain a complex feature representation, the gradient flow during the backward pass improves. Further,  our model has a negligible parameter overhead as there is a single scalar parameter associated with each layer in the network.Therefore, we expect BERMo architecture to be better candidate for compression and hence, we believe the proposed model could be ideal for resource-constrained settings.
    
%\noindent \textcolor{blue}{Results and Contributions}
%%% Update next para depending on the results 
\par 
    \textbf{Contributions: } The contributions of this work can be summarized as follows:
    \begin{enumerate}[label=\roman*]
        \item \label{cont:standalone} We propose BERMo, which generates complex feature maps using linear combination of features from different depths. We evaluate the proposed model on the probing task from SentEval dataset (\cite{probing}) and find our model performs $2.67\%$ better than the baseline on the semantic tasks (Tense, Subjnum, Objnum, Somo, Coordinv)
        %% Need for reference?
        %% Sangamesh : I have referred the dataset in this line " SentEval dataset (\cite{probing}) "
        on average.
        \item \label{cont:stability} We observe our approach is stable when pruning with smaller datasets like SST-2 (\cite{glue}), where BERT commonly diverges.
        \item \label{cont:convergence}  We show our model supports higher pruning rates when compressing and converges $1.67\times$ and $1.15\times$  faster than BERT on MNLI and QQP (\cite{glue}),
        %% MNLI and QQP (reference)?
        %% Sangamesh Added.
        respectively.
        \item \label{cont:par_eff}  For loss penalty based pruning method our approach can obtain better parameter efficiency, $1.35\times$ for QQP, than BERT model for comparable performance.
        \item \label{cont:kd}  Our approach produces comparable results to the baseline for Knowledge Distillation with marginal improvements on SQuAD dataset.
    \end{enumerate}
    
    \begin{wrapfigure}{R}{0.62\textwidth}
        \centering
        \includegraphics[width=0.61\textwidth]{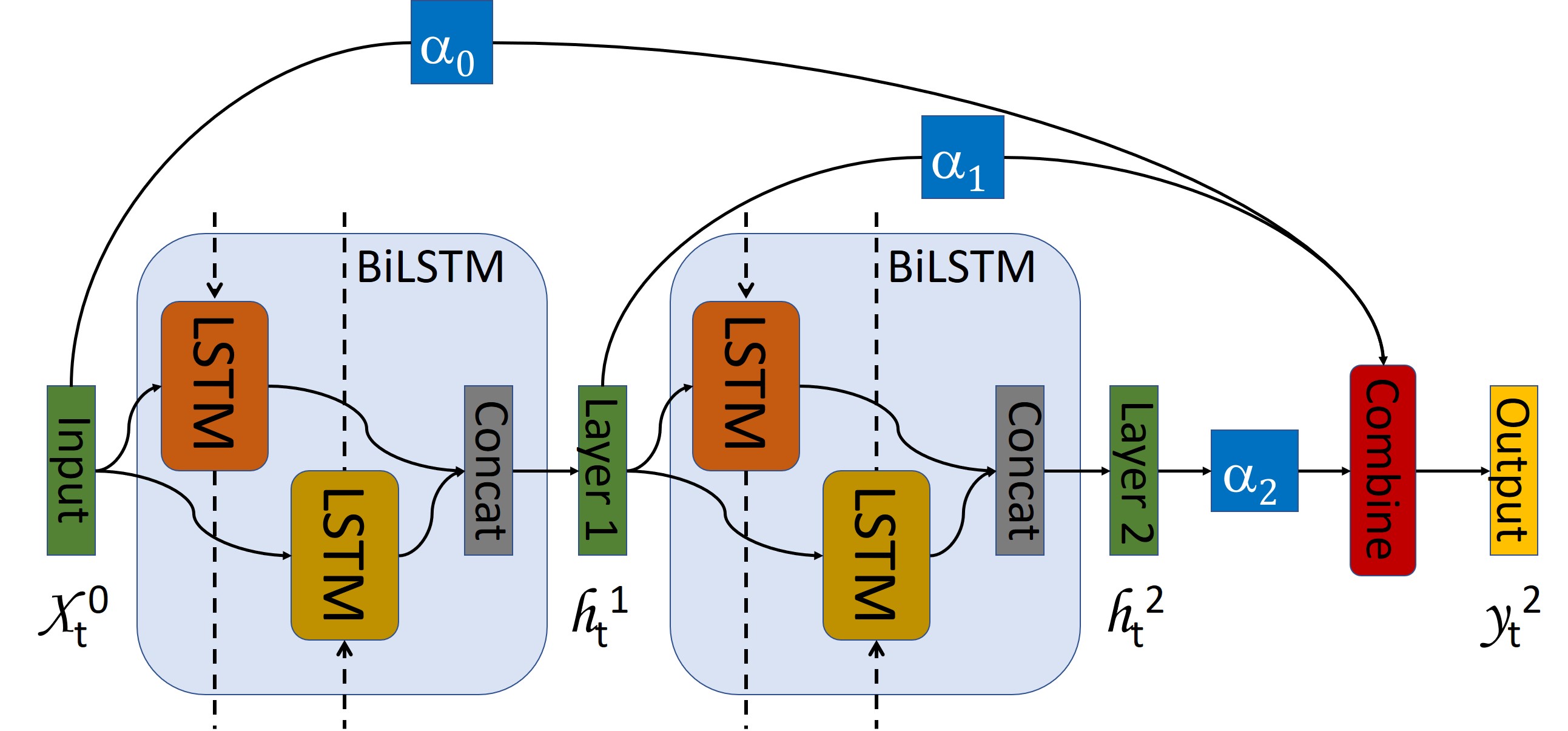}
        \caption{ELMo model Architecture \cite{elmo}}
        \label{fig:elmo}
        \begin{equation}
            \label{eqn:elmo}
            \mbox{\textbf{Combine}}= \gamma \sum_{j=0}^{L}  \alpha_{j} \times \mbox{\textbf{h}}_{j}
        \end{equation}
    where, $h_{j}$ is the activation, $\alpha_{j}$ is a learnt parameter associated with layer $j$ and $\gamma$ is the scaling factor learnt during training.
    \vspace{-35pt}
    \end{wrapfigure}
    \textbf{Outline: }The rest of the paper is organised as follows: Section \ref{sec:background} describes ELMo (\cite{elmo}), BERT (\cite{bert}) and Pruning methods. Section \ref{sec:methodology} elaborates the proposed model. The experimental setup and the results are presented in Section \ref{sec:results}. In Section \ref{sec:discussion} we summarize our work and discuss the future possibilities. Section \ref{sec:related} reports the related work and we conclude our paper with Broader Impact in Section \ref{sec:impact}.
%\newpage
\section{Background}\label{sec:background}
\subsection{ELMo}
\par
    ELMo (\cite{elmo}) studies the features at different depths of a Bi-LSTM architecture. The authors analyze a two-layered Bi-LSTM network, where they illustrate that the first layer is good at capturing the syntactic aspects and the second layer captures the semantic information.  Further, the authors of this work propose an architecture shown in  Figure \ref{fig:elmo}, to combine the features from the different layers to obtain complex features representations. Equation (\ref{eqn:elmo}) presents the mathematical details for mixing the activations from different depths in this architecture.
    The authors present a two-stage training process. The first stage, called the pre-training phase, involves training the network on a large unlabelled corpus in an unsupervised fashion. 
    The second stage, popularly known as the fine-tuning phase, deals with supervised training on a downstream task. Due to the tremendous generalization improvements, practitioners have widely adopted this two-stage training methodology.
    
\subsection{BERT}
\par
    \begin{wrapfigure}{R}{0.62\textwidth}
        \centering
        \includegraphics[width=0.61\textwidth]{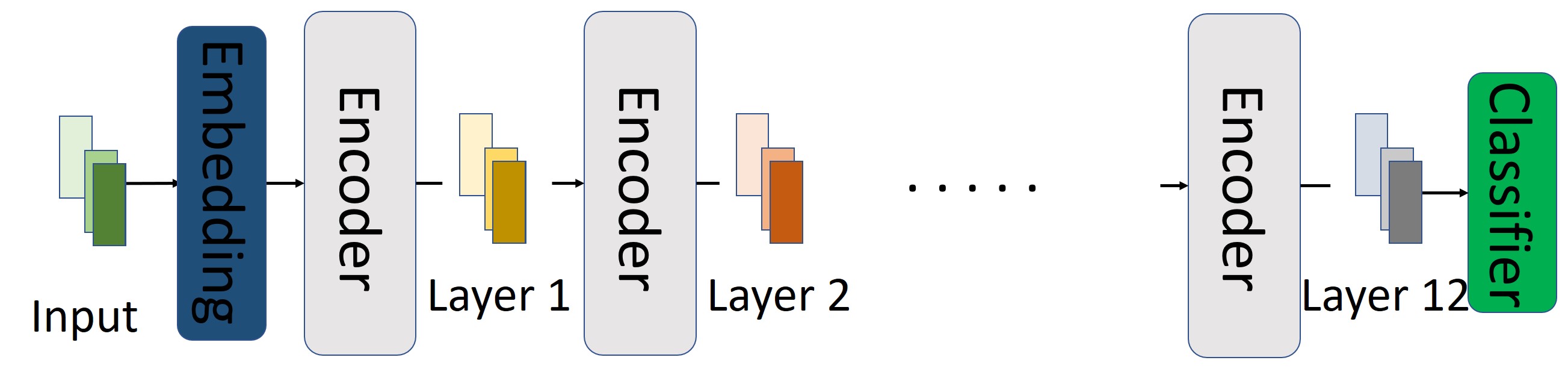}
        \caption{BERT model architecture \cite{bert}}
        \label{fig:bert}
    \end{wrapfigure}
    
    Similar to ELMo, BERT (see Figure \ref{fig:bert}) uses the two-stage training process on the encoder network from the Transformer model. The pre-training stage performs self-supervised learning on a large corpus of training data. The model is trained on Masked Language Modeling (MLM) and Next Sentence Prediction (NSP) objectives in the semi-supervised training phase. MLM objective trains the model to predict the masked, replaced, or retained words. Such a scheme enables bidirectional information flow for the prediction of a particular word. The NSP task is oriented for the model to learn the relationships between the sentences. The authors used two additional embeddings along with the token embedding: the segment embedding to distinguishing the different sentences, and the position embedding that marks the position of the words in the input. We refer the readers to (\cite{bert}) for implementation details.

\subsection{Pruning}
\par
    The general form of pruning method can be expressed as $a = (W \odot M)x$, where $x$ is the input to the current layer, $a$ is the activation of the output layer, $W$ is the layer weight, $M$ is the pruning mask, and $\odot$ is the Hadamard product (element-wise product). 
    The pruning mask $M$ is the same size as the weights, with each element commonly restricted to 0 or 1. $M$ is a function of the importance scores, $S$, of the weights, mathematically represented by $M=f_p(S)$, where $f_p(.)$ is some nonlinear function. Any pruning approach can be defined by the algorithm used to compute the importance score (S) and the mapping function ($f_p(.)$) used to compute the mask $M$.
    %Different pruning algorithms differ in the way they realise these importance scores and the function mapping the importance score to the $M$.
    %% The above sentence does not read well. I am not sure if it's grammatically correct the way I am reading it. Please update.
    %%Sangamesh: Sentence rewritten. 
    
    \subsubsection{Magnitude Pruning}\label{subsec:mag_prune}
    One of the well known approaches, Magnitude Pruning (\cite{han2015deep}), defines $S_{i,j} = (|W_{i,j}|)$ or the magnitude of the weights and the function $f_p(.) =$ Top\textsubscript{v}$(.)$ given by equation (\ref{eq:topv}).  This algorithm is 0\textsuperscript{th} order as it does not rely on the gradients of the weights.
    \begin{equation}
    \label{eq:topv}
        \text{Top\textsubscript{v}}(S_{i,j}) = \left\{ \begin{array}{cc} 
                        1 & \hspace{5mm} S_{i,j} \text{ in top v \%} \\
                        0 & \hspace{5mm} \text{otherwise} \\
                        \end{array} \right.    
    \end{equation}
    \subsubsection{$L_{0}$ Pruning}\label{subsec:l0_prune}
    This approach uses gradient information to compute $S$. This is a stochastic pruning approach as it samples a parameter from uniform distribution. The mask function $f_p(.)$ is given by equation (\ref{eq:continuous-hard-concrete1}, \ref{eq:continuous-hard-concrete2}) and the Scores (S) are computed using equation (\ref{eq:l0_score}). 
    At the time of inference, the non stochastic version, given by
    %% What is the non stochastic version -- it appears all of a sudden
    %% Sangamesh: I added "This is a stochastic pruning approach as it samples a parameter from a uniform distribution." in the begining.  
    equation (\ref{eq:non_stochastic_mask}) is used. In this approach L\textsubscript{0} penalty is added to the optimization which indirectly controls the levels of sparsity in the model.

    \begin{equation}
    \label{eq:continuous-hard-concrete1} 
       u\sim\mathcal{U}(0,1), \qquad  \overline{S}_{i,j} = \sigma((log(u) - log(1-u) + log(S_{i,j}))/\beta)
    \end{equation}
    
    \begin{equation}
    \label{eq:continuous-hard-concrete2} 
       Z_{i,j} = \overline{S}_{i,j}  (\epsilon -\gamma) + \gamma  \quad M_{i,j} = \text{min}(1, \text{max}(0, Z_{i,j}))
    \end{equation}
    where, $\beta$, $\gamma$ and $\epsilon$ are hyperparameters.

    \begin{equation}
    \label{eq:l0_score}
    S_{i,j} = -\sum_{t} \frac{\partial \mathcal{L}}{\partial W_{i,j}} \times W_{i,j} \times g(\overline{S}_{i,j}),  \quad g(\overline{S}_{i,j}) = \frac{(\epsilon - \gamma) }{\beta} \times \overline{S}_{i,j} \times (1 - \overline{S}_{i,j} ) \times  1_{\{1\leq Z_{i,j} \leq 1\}}
    \end{equation}

    \begin{equation}
    \label{eq:non_stochastic_mask}  
    M = \text{min} (1, \text{max}(0, \frac{(\epsilon- \gamma)}{\beta} \sigma(S) + \gamma ) )
    \end{equation}

    \subsubsection{Movement Pruning} \label{subsec:mvp_prune}
    Movement pruning (\cite{movement-pruning}) gives importance to the changes in the weights during the fine-tuning task instead of just relying on the magnitude of the weight. The core idea is to prune the weights that shrink during the fine-tuning process. Such a scheme retains lower values of weights with an increasing magnitude over the higher weights with shrinking magnitudes.
    Mathematically, the masking function $f_p(.)$ is the same as the Magnitude pruning (equation (\ref{eq:topv})) and the Scores $S$ is given by equation (\ref{eq:mvp_score}). 
    %Here, we learn $S$ during the fine-tuning stage. 
    As the gradient of Top\textsubscript{v} is 0 wherever it is defined, we use the straight-through estimator (\cite{straight-through}).  In order to gradually prune the weights, cubic sparsity scheduler (\cite{zhu2017prune}) is introduced. 
    We follow the implementation from (\cite{movement-pruning})  
    \begin{equation}
    \label{eq:mvp_score}
        S_{i,j} = -\sum_{t} \frac{\partial \mathcal{L}}{\partial W_{i,j}} \times W_{i,j}
        %\text{Top\textsubscript{v}}(S_{i,j}) = \left\{ \begin{array}{cc} 
        %                1 & \hspace{5mm} S_{i,j} \text{ in top v \%} \\
        %                0 & \hspace{5mm} \text{otherwise} \\
        %                \end{array} \right.    
    \end{equation}

\subsection{Knowledge Distillation}
\par
    Knowledge Distillation (KD) (\cite{kd}) is a technique where a smaller model, called the student model, is trained to mimic the logits of a bigger network, called the teacher model. This technique is different from transfer learning as the student model is different from the teacher model, and the weights of the students are not initialized based on the teacher model. KD improves the generalization capacity of the smaller network (\cite{jiao2019tinybert,sanh2019distilbert}). In order to perform KD, its not necessary to have access to the training dataset. One can train the student network to mimic the teacher network on random inputs. 

%\newpage
\section{Methodology} \label{sec:methodology}
\par
    \subsection{BERMo}
    \begin{figure}[htbp]%{R}{1.00\textwidth}
        \centering
        \includegraphics[width=0.8\columnwidth]{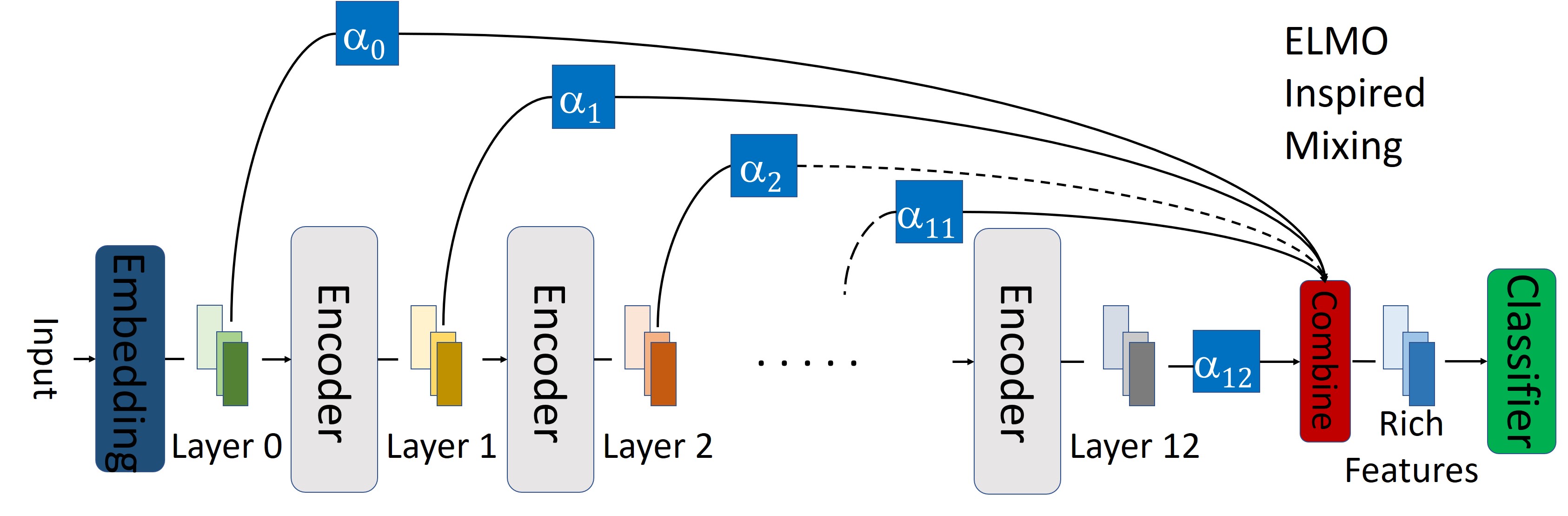}
        \caption{Proposed model architecture BERMo}
        \label{fig:elbert}
    \end{figure}
    \begin{wrapfigure}{R}{0.62\textwidth}
        \centering
        \includegraphics[width=0.61\textwidth]{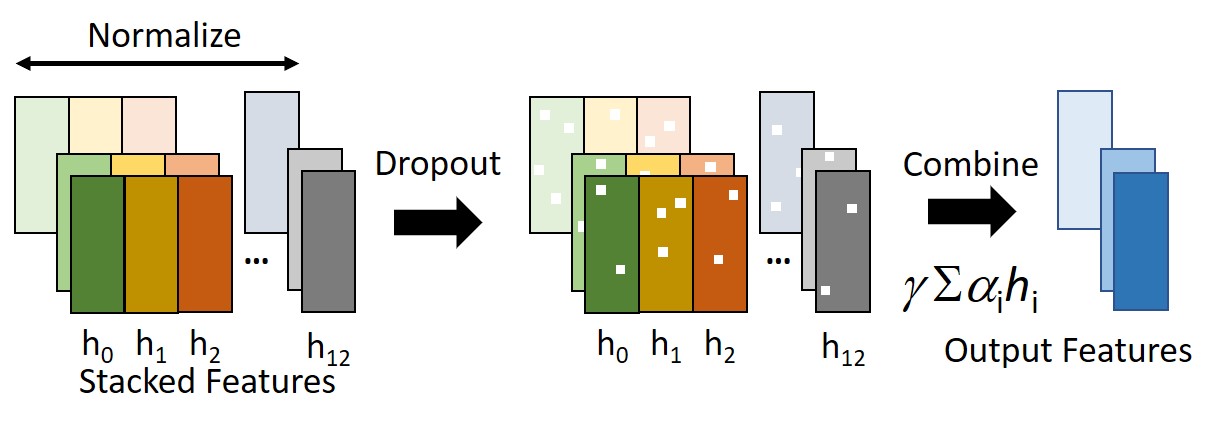}
        \caption{Combine Block}
        \label{fig:mixing}
    \end{wrapfigure}
    BERMo, shown in Figure \ref{fig:elbert}, modifies the BERT architecture by linearly combining the features using the scheme proposed in ELMo in order to obtain a rich feature representation. 
    Learnable scalar parameters $\alpha_{i}$ is associated with the activations of $i^{th}$ layer with $\alpha_0$ being the scalar parameter associated with embedding of the input. 
    These scalar parameters are softmax normalized to ensure $\sum_{i=0}^{12}\alpha_{i}=1$. 
    The \textbf{combine block} is used to obtain the weighted-average of the features from different layers and scales the result with a learnable parameter $\gamma$ following the equation (\ref{eqn:elmo}). 
    The parameters $\alpha$'s are initialized using Xavier initialization (\cite{glorot2010understanding}) and $\gamma$ is initialized to 1.

    To ensure the inputs are on similar scale, we normalize the stacked features across the layer as shown in Figure \ref{fig:mixing}. 
    This is conceptually similar to Layer Normalization (\cite{ba2016layer}) where channels are replaced by the features of the layers. 
    Dropout (\cite{srivastava2014dropout}) is used as a regularizer to avoid the dependence of neural network on specific neurons. 
    Finally, the features maps from each layer are combined with the scalar weights as per equation (\ref{eqn:elmo}) to obtain the weighted average of features from different depths. 
    %The code snippet for the combine block can be found in Appendix \ref{apx:combine_code}.
    
    % \subsection{Why skip connections}
    %%% next draft improvements

   % \subsection{Why ELMo skip connection}

%\newpage
%\input{Sections/experiment}
%\newpage
\section{Results}\label{sec:results}
    
     In our experiments we consider BERT-based-uncased\footnote{\url{https://huggingface.co/bert-base-uncased}} as our baseline model which roughly contains 84M parameters. Our experiments are built on top of Movement Pruning Repository by Huggingface (\cite{movement-pruning})\footnote{\url{https://github.com/huggingface/block_movement_pruning}}. We call this model as BERT\textsubscript{BASE} throughout the paper. We refer to the model with our approach applied to BERT\textsubscript{BASE} as BERMo\textsubscript{BASE}. 
     We run all our experiments on RTX 2080 Ti having Intel Xeon Silver 4114 CPU (2.20 GHz). For all the tasks we use a single card, except for SQuAD simulations where we use 2 cards to support longer sequence length. Detailed list of hyperparameters for all the experiments can be found in Appendix \ref{apx:hyperparameters}. We found that higher learning rates for the skip connections improved the performance of the model and hence the learning rate for the skip connections was kept same as the mask learning rate ($10^{-2}$). The code for the experiments in this work can be found on \url{https://anonymous.4open.science/r/BERMo-112D/README.md}
    
    In this section, we present the results for various experiment. We assess the performance of our model on different linguistic tasks (\ref{subsec:standalone}).  In Subsection \ref{subsec:pruning}, we test the fine-pruning performance of the models for stability (\ref{subsec:stability}), convergence (\ref{subsec:convergence}), different pruning approaches (\ref{subsec:diff_pruning}) and  knowledge distillation (\ref{subsec:knowledge_distillation}).
    
\subsection{Standalone Performance of BERMo} \label{subsec:standalone}
\par    

    %\begin{center}
        \begin{table}[htpb]
            \centering
        \resizebox{\columnwidth}{!}{
            \begin{tabular}{|c| c c | c c c | c c c c c|} 
                 \hline
                 \multirow{2}{*}{Model}& 
                 \multicolumn{2}{c|}{Surface}&
                 \multicolumn{3}{c|}{Syntactic}&
                 \multicolumn{5}{c|}{Semantic}\\
                 &Sentlen& 
                 WC & 
                 TreeDepth & 
                 TopConst & 
                 Bshift & 
                 Tense & 
                 SubjNum & 
                 ObjNum & 
                 Somo & 
                 CoordInv\\
                 \hline
                 \hline
                    BERT\textsubscript{BASE} &
                     $99.12$ & 
                     $\textbf{84.16}$&
                     $\textbf{72.77}$& 
                     $\textbf{77.07}$& 
                     $96.22$& 
                     $82.57$& 
                     $87.43$& 
                     $90.71$& 
                     $61.51$& 
                     $82.95$ \\
                    BERMo\textsubscript{BASE} &
                     $\textbf{99.24}$ & 
                     $83.39$ &
                     $72.36$ & 
                     $76.85$ & 
                     $\textbf{96.30}$ & 
                     $\textbf{87.22}$ & 
                     $\textbf{90.21}$ & 
                     $\textbf{93.68}$ & 
                     $\textbf{64.27}$ & 
                     $\textbf{83.16}$\\
                    \hline
                    Improvements &
                     $\textbf{0.036}$ & 
                     $-0.771$ &
                     $-0.41$ & 
                     $-0.214$ & 
                     $\textbf{0.082}$& 
                     $\textbf{4.646}$ & 
                     $\textbf{2.78}$ & 
                     $\textbf{2.968}$ & 
                     $\textbf{2.756}$ & 
                     $\textbf{0.212}$\\
                 \hline
            \end{tabular}
            }
            \caption{ 
                Accuracy comparison between BERT\textsubscript{BASE} and BERMo\textsubscript{BASE} on the probing task from the SentEval dataset. The results in the table correspond to the average accuracy over 5 runs. Two of the runs diverged for the WC tasks on the BERT\textsubscript{BASE} model and hence the results for WC are averaged over three runs.
            }
            \label{tab:standalone_probing}
        \end{table}
    %\end{center}

    As we combine the features from different depths to obtain a rich representation, we follow \cite{bert-language-structure} and evaluate our model on the probing tasks from the SentEval dataset (\cite{probing}). Tasks of this dataset are categorised into Surface (Sentence Length-Sentlen, Word content-WC), Sytactic (TreeDepth, TopConst, and Bshift) and Semantic tasks (Tense, SubjNum, ObjNum, somo and coordinv), helping us examine what linguistic tasks benefit from the BERMo architecture. Further, all the tasks of this dataset have 100k training, 10k validation and 10k testing samples with each class having identical number of samples. These properties of the dataset allow us to compare the raw accuracies of the model without worrying about the data imbalance issue. In this subsection, we evaluate BERMo\textsubscript{BASE} against BERT\textsubscript{BASE} on these probing task from the SentEval dataset to understand what linguistic feature benefit from our approach. 
    All the models in these experiments are run for 3 epochs with the batch size of 32 and a sequence length of 128 following (\cite{movement-pruning}). 
    Further, to ensure the results are statistically significant we run our experiment with 5 randomly chosen initialization. 
    The detailed results for each initialization are presented in Appendix \ref{apx:standalone_trails}. 
    Table \ref{tab:standalone_probing} presents the average over these 5 trails. We find our approach outperform BERT\textsubscript{BASE} on 7 of these 10 tasks. Semantic tasks benefit the most from complex feature representation enabled by our approach. All the five semantic tasks show improvements in accuracy with the maximum improvement of $\sim 4.646 \%$ obtained on Tense task and an average improvement of $\sim 2.673 \%$ (Contribution (\ref{cont:standalone})).  
    We find our approach improves over the baseline by $\sim 1.21\%$ across all the 10 tasks on average.

\subsection{Pruning} \label{subsec:pruning}
\par

    %\begin{center}
        \begin{table}[t]
            \centering
            \begin{tabular}{|c|c|c|c|c|} 
                 \hline
                 \multirow{2}{*}{Model} & 
                 SST-2 &
                 MNLI &
                 QQP&
                 SQuAD\\
                 & Acc & ACC-m / ACC-mm & Acc./F1 & F1\\
                 \hline
                 \hline
                    BERT\textsubscript{BASE} &
                     $91.74$&
                     $\textbf{80.58/81.2}$&
                     $\textbf{88.62/84.53}$& 
                     $76.95$\\
                    BERMo\textsubscript{BASE} &
                     $\textbf{92.43}$&
                     $80.04/80.84$ &
                     $88.27/84.29$ & 
                     $\textbf{77.43}$  \\
                 \hline
            \end{tabular}
            \caption{Accuracy/F1-Score comparison results for SST-2, MNLI, QQP and SQuAD for BERT\textsubscript{BASE} and BERMo\textsubscript{BASE}.}
            \label{tab:standalone_pruning}
        \end{table}
    %\end{center}

    As we introduce skip connections in our model, we expect our model to train in a stable and efficient manner when subject to pruning techniques. 
    In our experiments, we find pruning approach diverges on some of the probing tasks for the baseline model. 
    We believe this unstable pruning behaviour is due to the small size of the dataset. 
    We test this hypothesis with Stanford Sentiment Treebank-2 (SST-2) task from the General Language Evaluation and Understanding (GLUE) dataset (\cite{glue}) and find the baseline model diverges most of the times whereas our approach is compatible with smaller datasets. 
    To fairly compare our approach with BERT\textsubscript{BASE}, for the pruning task we evaluate our approach on larger datasets like Multi-genre Natural Language Inference (MNLI) and Quora Questions Pair (QQP) tasks from the
    %% Why therefore you use MNLI and QQP? Therefore part does not seem to clearly state why?
    %% Sangamesh: I added few lines to explain " We believe this unstable pruning behaviour is due to the small size of the dataset. We test this hypothesis with Stanford Sentiment Treebank-2 (SST-2) task from the General Language Evaluation and Understanding (GLUE) dataset (\cite{glue}) and find the baseline model diverges most of the times whereas our approach is compatible with smaller datasets. To fairly compare our approach with BERT\textsubscript{BASE},"
    GLUE dataset and Stanford Question Answering (SQuAD) dataset (\cite{squad1.1}). 
    For comparing the model performance we use the F1 score for SQuAD, accuracy and F1 score for QQP, accuracy for MNLI and accuracy for SST-2.
    These datasets are selected in order to be comparable to \cite{movement-pruning}. 
    The performance of our model on these tasks, without the compression, are reported in Table \ref{tab:standalone_pruning}. The hyperparameters for this experiment are kept same as \cite{movement-pruning} and are reported in Appendix \ref{apx:hyperparameters}.  
    
    For our analysis in Subsection (\ref{subsec:stability}, \ref{subsec:convergence}, \ref{subsec:knowledge_distillation} ), we chose Movement Pruning (\ref{subsec:mvp_prune}) (\cite{movement-pruning})  for benchmarking our approach against the baseline. 
    This approach was chosen as it allows us to control the exact pruning fraction in the experiments, which allows us to fairly compare different models. We compare the two approaches for nearly $90\%$ compression in these sections. 
    Further in Subsection (\ref{subsec:diff_pruning}), we maintain $\sim10\%$ weight retention for magnitude and movement pruning approaches, however, for penalty based approaches we select the hyperparameter to be close to $\sim 90\%$ pruning from \cite{movement-pruning} for BERT\textsubscript{BASE}. The hyperparameters for BERMo\textsubscript{BASE} for loss penalty based methods were selected to match the performance of the baseline.

\subsubsection{Stability under Pruning} \label{subsec:stability}
%\begin{center}
        \begin{table}[htpb]
            \centering
            \resizebox{\columnwidth}{!}{
            \begin{tabular}{|c|c|c|c|c|c|c|c|c|c|c|} 
                 \hline
                 \multirow{2}{*}{Model} & 
                 %Remaining&
                 Seed &
                 Seed &
                 Seed &
                 Seed &
                 Seed &
                 Seed &
                 Seed &
                 Seed &
                 Seed &
                 Seed \\
                 %& Weights
                 & 9 & 25& 39 & 52& 59&  63 & 77 & 87& 91& 96\\
                 \hline
                 \hline
                    BERT\textsubscript{BASE} &
                     %$10\%$& 
                     $79.85$&
                     Diverges& 
                     Diverges& 
                     $82.22$& 
                     Diverges& 
                     Diverges& 
                     Diverges& 
                     Diverges& 
                     Diverges& 
                     Diverges\\
                    BERMo\textsubscript{BASE} &
                     %$10\%$& 
                     $87.39$ &
                     $87.61$& 
                     $85.88$ & 
                     $87.04$& 
                     $87.39$& 
                     $86.47$ & 
                     $87.39$ & 
                     $86.70$& 
                     $86.58$& 
                     $86.24$\\
                 \hline
        \end{tabular}}
        \caption{Stability comparison on SST-2 dataset for different initialisation.}
        \label{tab:stability}
    \end{table}
%\end{center}
    During our experiment with the probing task we found the baseline approach to diverge when we used movement pruning to retain $10\%$ of the weights. Such divergence was not seen when pruning was applied to our model. These divergence issues were significant when the training dataset size is small. The skip connection in our model leads to better gradient flow and we believe that improved gradient flow is important for the stability of the pruning. To verify this claim, we evaluate the proposed model for pruning on SST-2 which is a small dataset having 67k training samples. 
    The results in Table \ref{tab:stability} show that the BERT\textsubscript{BASE} model in most cases diverges with this dataset. We also found that the BERT\textsubscript{BASE} model diverged for two of the seeds for Word Content task, showing that our approach can also stabilize the fine-tuning phase. 
    Our proposed model did not diverge during the pruning or the training process in any of our experiments (Contribution (\ref{cont:stability})).

\subsubsection{Convergence}\label{subsec:convergence}
    We analyze the convergence by monitoring how the performance of the models varies with the change in epoch. 
    As fine-pruning with smaller dataset was challenging for the baseline approach, we chose to evaluate the convergence for MNLI, QQP and SQuAD tasks. For this study we employ movement pruning to retain $\sim 10\%$ of the model weights. We use the cubic sparsity scheduler (\cite{zhu2017prune}), where the model gradually reduces the network weights over the epochs, pruning certain amount of weights as dictated by the sparsity scheduler. Hence, varying the number of epochs also controls the speed of pruning i.e. lower epochs means more parameters being pruned per epoch to attain the same desired sparsity level ($\sim 10\%$) at the end the fine-pruning.
    
\begin{figure}[htbp]%{R}{1.00\textwidth}
        \centering
        \includegraphics[width=0.85\columnwidth]{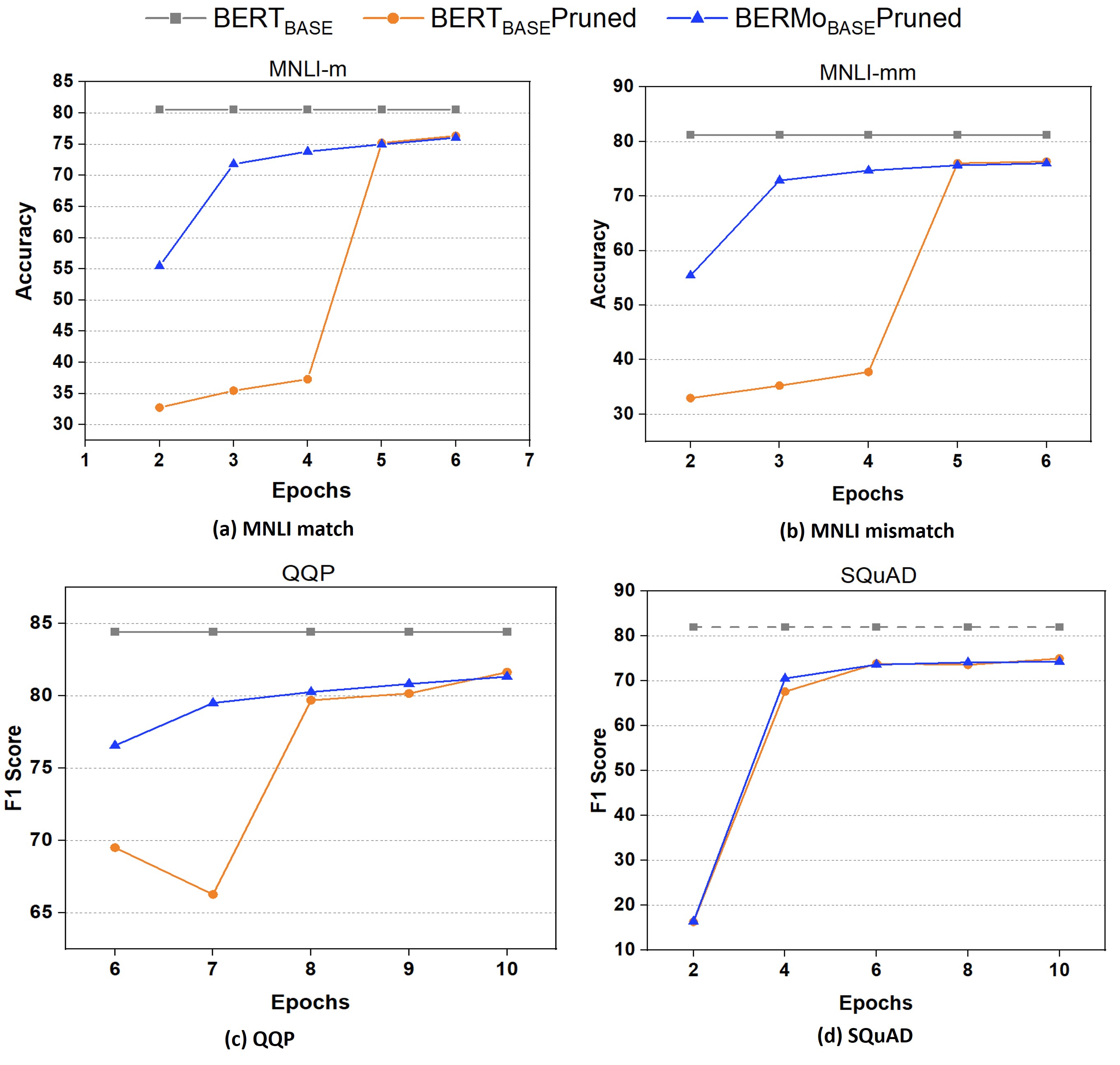}
        \caption{Accuracy/F1-Score comparison with varying epochs for pruning BERT\textsubscript{BASE} and BERMo\textsubscript{BASE} to retain 10\% of the weights. BERT Baseline represents the performance of the model trained for 3 epochs without any pruning (Table \ref{tab:standalone_pruning}).  }
        \label{fig:convergence}
    \end{figure}

    Figure \ref{fig:convergence} shows the results for MNLI, QQP and SQuAD. We see that our approach converges significantly faster on MNLI ($1.67\times$) and QQP ($1.15 \times$) while the convergence on SQuAD is similar to the baseline (Contribution (\ref{cont:convergence})). On our setup, each epoch for MNLI and QQP took roughtly 5hrs. The faster convergence obtained by our method hence saves 10hrs and 5hrs on MNLI and QQP respectively.
    Moreover, less epochs also means that the model supports higher pruning rates as the model is able to prune $90\%$ of weights is lesser epochs. 
\subsubsection{Different Pruning} \label{subsec:diff_pruning}
\par
    We compare our model with Magnitude Pruning (\ref{subsec:mag_prune}), L\textsubscript{0} Pruning (\ref{subsec:l0_prune}), Movement Pruning and Soft Movement Pruning (\ref{subsec:mvp_prune}). We train the both the models with the best hyperparameters (\cite{movement-pruning}) for the baseline to ensure that the baseline converges. We find our model performance is similar to the baseline. The pruning threshold for Magnitude and Movement pruning approaches are set to retain $\sim 10\%$ of the weights. The pruning percentage for the loss penalty based approaches like the L\textsubscript{0} and Soft Movement Pruning can be tuned by varying the Lagrange multiplier ($\lambda$) for the pruning loss term. However, it is difficult to exactly control the percentage of weights retain with these approaches. Hence, we train the BERT\textsubscript{BASE} model with the hyperparameters suggested by (\cite{movement-pruning}) and for experiment on our model we select lambda such that the performance of our model matches the baseline model performance.  
    
    The result of these experiments is presented in Table \ref{tab:different_pruning}. Although the model is trained with the optimal hyperparameters for the baseline our model achieves comparable results to BERT\textsubscript{BASE} for the Magnitude and Movement pruning approach. 
    Surprisingly, for penalty based methods we are able to prune more aggressively for comparable accuracies on QQP task (Contribution (\ref{cont:par_eff})). 
    In the case of L\textsubscript{0} regularization, our model has $1.4 \times$ less weights as compared to BERT\textsubscript{BASE} for QQP. 
    Similarly, for soft movement pruning approach our model has $1.29\times$ less weight as compared to BERT\textsubscript{BASE} for QQP.
    Soft movement pruning is the State-of-The-Art approach in literature (\cite{movement-pruning}) and the results in Table \ref{tab:different_pruning} show our model can potentially prune more weights for comparable accuracy for some tasks. 
    
    %\begin{center}
        \begin{table}[htpb]
            \centering
             \resizebox{\columnwidth}{!}{
            \begin{tabular}{|c|c|c|c|c|c|c|c|} 
                 \hline
                 \multirow{3}{*}{Model} & 
                 &
                 \multicolumn{2}{c|}{MNLI} &
                 \multicolumn{2}{c|}{QQP}&
                 \multicolumn{2}{c|}{SQuAD}\\
                 \cline{3-8}
                 
                 & Pruning
                 & Remaining  
                 & 
                 & Remaining  
                 & 
                 &  Remaining  
                 & \\
                  
                 & Approach
                 & Weights
                 & ACC-m / ACC-mm
                 & Weights
                 & Acc./F1 
                 & Weights
                 &  F1 \\%&  &
                 \hline
                 \hline
                    BERT\textsubscript{BASE} &
                     \multirow{2}{*}{Magnitude}&
                     \multirow{2}{*}{$10.1\%$}& 
                     $74.9/76.05$ & 
                     \multirow{2}{*}{$10.1\%$}&
                     $86.37/81.41$ & 
                     \multirow{2}{*}{$10.1\%$}&
                     $71.05$\\
                    BERMo\textsubscript{BASE} &
                     & 
                     &
                     $74.2/75.19$ & 
                     &
                     $85.81/80.97$ & 
                     &
                     $70.93$\\
                 \hline
                    BERT\textsubscript{BASE} &
                     \multirow{2}{*}{L0}&
                     $7.6\%$& 
                     $74.42/74.96$ & 
                     $7.7\%$& 
                     $85.85/79.44$ & 
                     $9.4\%$& 
                     $72.72$\\
                    BERMo\textsubscript{BASE} &
                     & 
                     $7.5\%$&
                     $74.27/74.91$ & 
                     $5.5\%$& 
                     $85.78/80.67$ & 
                     $9.5\%$& 
                     $72.46$\\
                 \hline
                    BERT\textsubscript{BASE} &
                     \multirow{2}{*}{Movement}&
                     \multirow{2}{*}{$10.1\%$}& 
                     $76.32/76.67$& 
                     \multirow{2}{*}{$10.1\%$}& 
                     $86.28/81.64$&
                     \multirow{2}{*}{$10.1\%$}& 
                     $74.99$\\
                    BERMo\textsubscript{BASE} &
                     & 
                      &
                     $76.03/76.32$ & 
                     &
                     $86.02/81.33$ & 
                     &
                     $74.28$\\
                 \hline
                    BERT\textsubscript{BASE} &
                     Soft&
                     $12.2\%$& 
                     $77.83/78.39$& 
                     $14.9\%$& 
                     $87.6/83.23$&
                     $8.4\%$& 
                     $74.57$\\
                    BERMo\textsubscript{BASE} &
                     Movement&
                     $12\%$& 
                     $77.92/78.24$ & 
                     $11.5\%$& 
                     $87.49/83.21$ & 
                     $8.5\%$& 
                     $74.87$\\
                 \hline
            \end{tabular}}
            \caption{Comparing accuracy/F1-score of BERT\textsubscript{BASE} and BERMo\textsubscript{BASE} for different pruning approaches }
            \label{tab:different_pruning}
        \end{table}
    %\end{center}

\subsubsection{Knowledge Distillation} \label{subsec:knowledge_distillation}
\par
    Many approaches complement the pruning strategies with the knowledge distillation to recover the loss in performance caused by pruning. We also show improvement in the performance by Knowledge Distillation, see Table \ref{tab:knowledge_distillation}. Our approach gives better results on SQuAD and comparable results  for MNLI and QQP as compared to baseline with the optimal hyperparameters for the baseline (Contribution (\ref{cont:kd})). 
   %\begin{center}
        \begin{table}[htpb]
            \centering
            \begin{tabular}{|c|c|c|c|c|} 
                 \hline
                 \multirow{2}{*}{Model} & 
                 Remaining&
                 MNLI &
                 QQP&
                 SQuAD\\
                 
                 & Weights& ACC-m / ACC-mm & Acc./F1 & F1\\
                 \hline
                 \hline
                    BERT\textsubscript{BASE} &
                     $10.1\%$& 
                     $76.63/\textbf{77.46}$&
                     $81.79/\textbf{86.58}$& 
                     $76.69$\\
                    BERMo\textsubscript{BASE} &
                     $10.1\%$& 
                     $\textbf{76.82}/77.31$ &
                     $\textbf{81.86}/86.53$ & 
                     $\textbf{76.85}$  \\
                 \hline
            \end{tabular}
            \caption{Knowledge Distillation performance of BERT\textsubscript{BASE} and BERMo\textsubscript{BASE}}
            \label{tab:knowledge_distillation}
        \end{table}
    %\end{center}

%\newpage
\section{Discussion}\label{sec:discussion}
    \textbf{Summary: } This work presents BERMo, a model architecture which uses the features from different layers to obtain a complex feature representation. Our work shows that the proposed model performs better than the baseline on the semantic tasks. 
    %% imporvement over what?
    %% Sangamesh: updated
    Further, we find that the proposed model is more stable and converges faster (supports higher pruning rates) when subject to the compression techniques. 
    We find our approach gives better parameter efficiency than the baseline on the QQP task for penalty based pruning approaches.
    Moreover, our approach is comparable to BERT\textsubscript{BASE} when tested with Knowledge Distillation. 
    
    \textbf{Limitations: } The skip connections introduced in the network require the activation maps from different layers to be stored which marginally increases the memory requirement at the time of inference. The reduction in the training time and stability of the training compensate for this limitation.  
    
    \textbf{Future Directions: } We choose a simple weighted linear combination scheme, as the premise was to test the applicability of our model in the resource constrained setup. Our work paves a way for exploring more complex skip connection having more parameters in the quest to improve the baseline performance. 
    Further, we mix the features only in the last layer. We believe that such a summarization of features at each layer would benefit the model even further. Testing this approach would require pretraining the model from scratch as this scheme completely changes the activation maps at the input of each layer. 
    Although, we evaluate our model against BERT\textsubscript{BASE}, our approach could be extended to any Transformer based model as our methodology is agnostic to the choice of the backbone network. Moreover, as our method was able to converge in a fast and stable manner for pruning, we believe these trends should also hold during the pretraining phase. However, this needs to be extensively tested. 
%\subsubsection{Magnitude Pruning}
%\newpage
\section{Related Work}\label{sec:related}
\subsection{Linear combination of features from different layers in BERT}
    Linear combination scheme of ELMo has been used to quantify where the linguistic information is captured within the network (\cite{tenney2019bert}). 
    The authors of this work call this approach scalar mixing weights and analyze these scalar weight to interpret the linguistic contribution of a particular layer. 
    Concurrently, author's of (\cite{kondratyuk201975}) call these skip connections as Layer Attention and propose a model capable of accurately predicting universal part of speech, morphological features, lemmas, and dependency trees across 75 languages. 
    The architectural modification presented in this work is similar to (\cite{tenney2019bert, kondratyuk201975}), however, we intend to improve the baseline accuracies and enable faster and stable compression.
    In (\cite{su2020sesamebert}), the authors use global average pooling followed by two fully connected layers to obtain the scalar parameter associated with the network. The scale value is then used to compute the weighted average of the features from different layers. 
    Although, we target improvement in the baseline performance in our work our approach is computationally simpler and requires fewer parameters.
    
\subsection{Compression}
\par
    Weight Pruning (\cite{reed1993pruning,liang2021pruning}), knowledge distillation (\cite{gou2021knowledge}), weight quantization (\cite{gholami2021survey}), low rank matrix factorization (\cite{nguyen2019low}) are some approaches used to obtain the required level of compression. 
    Several research (\cite{sajjad2020poor, gordon2020compressing}) works tackle the issue by pruning  the parameters of the network. 
    In (\cite{movement-pruning}) the authors show using the gradient momentum information to select the pruning weights is superior to its counterpart which rely on the magnitude of weights.
    Certain works use quantization (\cite{zafrir2019q8bert,shen2020q,bai2020binarybert}) as a resort to compress the network. 
    Knowledge Distillation (KD) \cite{kd} is leveraged in some works (\cite{jiao2019tinybert,sanh2019distilbert}) to bridge the gap between a compact model and BERT.
    Further, many works \cite{mao2020ladabert,tukan2020compressed} use Low rank matrix factorization to deal with the issue. 
    The authors of ROSITA (\cite{rosita}) outline a methodology to combine weight pruning, KD and low rank factorization.
    Our work is orthogonal to these as it makes architectural modification to the standard architecture to support high pruning rates and improve the training stability for the network.
    
\iffalse
github links
https://github.com/kmjohnson/twitter-framing
\fi
\section{Broader Impact}\label{sec:impact}
    Research presented in this work improves the baseline performance on semantic tasks. 
    This work highlights the importance of adding skip connections to the network in improving the training convergence and stability. 
    We believe this work would act as a stepping stone and motivate further research in this direction, reducing the training time for these models. 
    These improved training speeds also make room for enlarging the dataset size generally correlated with improvements in generalization performance.
    Further, as this work deals with reducing the training time for pruning, a possible application would be online pruning on resource constrained setup.
    Moreover, from an environmental perspective reducing training time will reduce the carbon footprint of these large language models.

\section{Acknowledgement}
\par
    We acknowledge the authors of Movement Pruning (\cite{movement-pruning}) for their open source repository and timely response to our queries. 
    %The technical discussion with Timur Ibrayev, Deepak Ravikumar, Aparna Aketi and Amogh Agrawal, from Nano(neuro)electronic Research Laboratory (NRL) at Purdue University, is greatly appreciated. 
    Further, we would like to thank Deepak Ravikumar, Gobinda Saha, Nitin Rathi, Timur Ibrayev, Deepika Sharma and Utkarsh Saxena from Nano(neuro)electronic Research Laboratory (NRL) at Purdue University for their feedback to improve the write up.
    Finally, we thank the Center for Brain-Inspired Computing (C-BRIC), one of six centers in JUMP, funded by Semiconductor Research Corporation (SRC) and DARPA for supporting this work.

%\newpage
\bibliography{iclr2022_conference}

\begin{thebibliography}{42}
\providecommand{\natexlab}[1]{#1}
\providecommand{\url}[1]{\texttt{#1}}
\expandafter\ifx\csname urlstyle\endcsname\relax
  \providecommand{\doi}[1]{doi: #1}\else
  \providecommand{\doi}{doi: \begingroup \urlstyle{rm}\Url}\fi

\bibitem[Ba et~al.(2016)Ba, Kiros, and Hinton]{ba2016layer}
Jimmy~Lei Ba, Jamie~Ryan Kiros, and Geoffrey~E. Hinton.
\newblock Layer normalization, 2016.

\bibitem[Bai et~al.(2020)Bai, Zhang, Hou, Shang, Jin, Jiang, Liu, Lyu, and
  King]{bai2020binarybert}
Haoli Bai, Wei Zhang, Lu~Hou, Lifeng Shang, Jing Jin, Xin Jiang, Qun Liu,
  Michael Lyu, and Irwin King.
\newblock Binarybert: Pushing the limit of bert quantization.
\newblock \emph{arXiv preprint arXiv:2012.15701}, 2020.

\bibitem[Bengio et~al.(2013)Bengio, L{\'e}onard, and
  Courville]{straight-through}
Yoshua Bengio, Nicholas L{\'e}onard, and Aaron Courville.
\newblock Estimating or propagating gradients through stochastic neurons for
  conditional computation.
\newblock \emph{arXiv preprint arXiv:1308.3432}, 2013.

\bibitem[Brown et~al.(2020)Brown, Mann, Ryder, Subbiah, Kaplan, Dhariwal,
  Neelakantan, Shyam, Sastry, Askell, et~al.]{gpt}
Tom~B Brown, Benjamin Mann, Nick Ryder, Melanie Subbiah, Jared Kaplan, Prafulla
  Dhariwal, Arvind Neelakantan, Pranav Shyam, Girish Sastry, Amanda Askell,
  et~al.
\newblock Language models are few-shot learners.
\newblock \emph{arXiv preprint arXiv:2005.14165}, 2020.

\bibitem[Carion et~al.(2020)Carion, Massa, Synnaeve, Usunier, Kirillov, and
  Zagoruyko]{10.1007/978-3-030-58452-8_13}
Nicolas Carion, Francisco Massa, Gabriel Synnaeve, Nicolas Usunier, Alexander
  Kirillov, and Sergey Zagoruyko.
\newblock End-to-end object detection with transformers.
\newblock In Andrea Vedaldi, Horst Bischof, Thomas Brox, and Jan-Michael Frahm
  (eds.), \emph{Computer Vision -- ECCV 2020}, Cham, 2020. Springer
  International Publishing.

\bibitem[Conneau et~al.(2018)Conneau, Kruszewski, Lample, Barrault, and
  Baroni]{probing}
Alexis Conneau, German Kruszewski, Guillaume Lample, Lo{\"\i}c Barrault, and
  Marco Baroni.
\newblock What you can cram into a single vector: Probing sentence embeddings
  for linguistic properties.
\newblock \emph{arXiv preprint arXiv:1805.01070}, 2018.

\bibitem[Devlin et~al.(2019)Devlin, Chang, Lee, and Toutanova]{bert}
Jacob Devlin, Ming-Wei Chang, Kenton Lee, and Kristina Toutanova.
\newblock {BERT}: Pre-training of deep bidirectional transformers for language
  understanding.
\newblock In \emph{Proceedings of the 2019 Conference of the North {A}merican
  Chapter of the Association for Computational Linguistics: Human Language
  Technologies, Volume 1 (Long and Short Papers)}, pp.\  4171--4186,
  Minneapolis, Minnesota, June 2019. Association for Computational Linguistics.
\newblock \doi{10.18653/v1/N19-1423}.
\newblock URL \url{https://www.aclweb.org/anthology/N19-1423}.

\bibitem[Dosovitskiy et~al.(2021)Dosovitskiy, Beyer, Kolesnikov, Weissenborn,
  Zhai, Unterthiner, Dehghani, Minderer, Heigold, Gelly, Uszkoreit, and
  Houlsby]{dosovitskiy2021image}
Alexey Dosovitskiy, Lucas Beyer, Alexander Kolesnikov, Dirk Weissenborn,
  Xiaohua Zhai, Thomas Unterthiner, Mostafa Dehghani, Matthias Minderer, Georg
  Heigold, Sylvain Gelly, Jakob Uszkoreit, and Neil Houlsby.
\newblock An image is worth 16x16 words: Transformers for image recognition at
  scale, 2021.

\bibitem[Gholami et~al.(2021)Gholami, Kim, Dong, Yao, Mahoney, and
  Keutzer]{gholami2021survey}
Amir Gholami, Sehoon Kim, Zhen Dong, Zhewei Yao, Michael~W Mahoney, and Kurt
  Keutzer.
\newblock A survey of quantization methods for efficient neural network
  inference.
\newblock \emph{arXiv preprint arXiv:2103.13630}, 2021.

\bibitem[Glorot \& Bengio(2010)Glorot and Bengio]{glorot2010understanding}
Xavier Glorot and Yoshua Bengio.
\newblock Understanding the difficulty of training deep feedforward neural
  networks.
\newblock In \emph{Proceedings of the thirteenth international conference on
  artificial intelligence and statistics}, pp.\  249--256. JMLR Workshop and
  Conference Proceedings, 2010.

\bibitem[Gordon et~al.(2020)Gordon, Duh, and Andrews]{gordon2020compressing}
Mitchell~A Gordon, Kevin Duh, and Nicholas Andrews.
\newblock Compressing bert: Studying the effects of weight pruning on transfer
  learning.
\newblock \emph{arXiv preprint arXiv:2002.08307}, 2020.

\bibitem[Gou et~al.(2021)Gou, Yu, Maybank, and Tao]{gou2021knowledge}
Jianping Gou, Baosheng Yu, Stephen~J Maybank, and Dacheng Tao.
\newblock Knowledge distillation: A survey.
\newblock \emph{International Journal of Computer Vision}, 129\penalty0
  (6):\penalty0 1789--1819, 2021.

\bibitem[Han et~al.(2015)Han, Mao, and Dally]{han2015deep}
Song Han, Huizi Mao, and William~J Dally.
\newblock Deep compression: Compressing deep neural networks with pruning,
  trained quantization and huffman coding.
\newblock \emph{arXiv preprint arXiv:1510.00149}, 2015.

\bibitem[He et~al.(2016)He, Zhang, Ren, and Sun]{resnet}
Kaiming He, Xiangyu Zhang, Shaoqing Ren, and Jian Sun.
\newblock Deep residual learning for image recognition.
\newblock In \emph{Proceedings of the IEEE conference on computer vision and
  pattern recognition}, pp.\  770--778, 2016.

\bibitem[Hinton et~al.(2015)Hinton, Vinyals, and Dean]{kd}
Geoffrey Hinton, Oriol Vinyals, and Jeff Dean.
\newblock Distilling the knowledge in a neural network.
\newblock \emph{arXiv preprint arXiv:1503.02531}, 2015.

\bibitem[Huang et~al.(2016)Huang, Liu, and
  Weinberger]{DBLP:journals/corr/HuangLW16a}
Gao Huang, Zhuang Liu, and Kilian~Q. Weinberger.
\newblock Densely connected convolutional networks.
\newblock \emph{CoRR}, abs/1608.06993, 2016.
\newblock URL \url{http://arxiv.org/abs/1608.06993}.

\bibitem[Jaegle et~al.(2021)Jaegle, Gimeno, Brock, Zisserman, Vinyals, and
  Carreira]{jaegle2021perceiver}
Andrew Jaegle, Felix Gimeno, Andrew Brock, Andrew Zisserman, Oriol Vinyals, and
  Joao Carreira.
\newblock Perceiver: General perception with iterative attention, 2021.

\bibitem[Jawahar et~al.(2019)Jawahar, Sagot, and
  Seddah]{bert-language-structure}
Ganesh Jawahar, Beno{\^\i}t Sagot, and Djam{\'e} Seddah.
\newblock What does {BERT} learn about the structure of language?
\newblock In \emph{Proceedings of the 57th Annual Meeting of the Association
  for Computational Linguistics}, pp.\  3651--3657, Florence, Italy, July 2019.
  Association for Computational Linguistics.
\newblock \doi{10.18653/v1/P19-1356}.
\newblock URL \url{https://www.aclweb.org/anthology/P19-1356}.

\bibitem[Jiao et~al.(2019)Jiao, Yin, Shang, Jiang, Chen, Li, Wang, and
  Liu]{jiao2019tinybert}
Xiaoqi Jiao, Yichun Yin, Lifeng Shang, Xin Jiang, Xiao Chen, Linlin Li, Fang
  Wang, and Qun Liu.
\newblock Tinybert: Distilling bert for natural language understanding.
\newblock \emph{arXiv preprint arXiv:1909.10351}, 2019.

\bibitem[Kondratyuk \& Straka(2019)Kondratyuk and Straka]{kondratyuk201975}
Dan Kondratyuk and Milan Straka.
\newblock 75 languages, 1 model: Parsing universal dependencies universally.
\newblock \emph{arXiv preprint arXiv:1904.02099}, 2019.

\bibitem[Liang et~al.(2021)Liang, Glossner, Wang, and Shi]{liang2021pruning}
Tailin Liang, John Glossner, Lei Wang, and Shaobo Shi.
\newblock Pruning and quantization for deep neural network acceleration: A
  survey.
\newblock \emph{arXiv preprint arXiv:2101.09671}, 2021.

\bibitem[Liu et~al.(2021)Liu, Lin, and Yuan]{rosita}
Yuanxin Liu, Zheng Lin, and Fengcheng Yuan.
\newblock Rosita: Refined bert compression with integrated techniques.
\newblock In \emph{Proceedings of the AAAI Conference on Artificial
  Intelligence}, volume~35, pp.\  8715--8722, 2021.

\bibitem[Mao et~al.(2020)Mao, Wang, Wu, Zhang, Wang, Yang, Zhang, Tong, and
  Bai]{mao2020ladabert}
Yihuan Mao, Yujing Wang, Chufan Wu, Chen Zhang, Yang Wang, Yaming Yang, Quanlu
  Zhang, Yunhai Tong, and Jing Bai.
\newblock Ladabert: Lightweight adaptation of bert through hybrid model
  compression.
\newblock \emph{arXiv preprint arXiv:2004.04124}, 2020.

\bibitem[Nguyen et~al.(2019)Nguyen, Kim, and Shim]{nguyen2019low}
Luong~Trung Nguyen, Junhan Kim, and Byonghyo Shim.
\newblock Low-rank matrix completion: A contemporary survey.
\newblock \emph{IEEE Access}, 7:\penalty0 94215--94237, 2019.

\bibitem[Peters et~al.(2018)Peters, Neumann, Iyyer, Gardner, Clark, Lee, and
  Zettlemoyer]{elmo}
Matthew~E. Peters, Mark Neumann, Mohit Iyyer, Matt Gardner, Christopher Clark,
  Kenton Lee, and Luke Zettlemoyer.
\newblock Deep contextualized word representations.
\newblock In \emph{Proceedings of the 2018 Conference of the North {A}merican
  Chapter of the Association for Computational Linguistics: Human Language
  Technologies, Volume 1 (Long Papers)}, pp.\  2227--2237, New Orleans,
  Louisiana, June 2018. Association for Computational Linguistics.
\newblock \doi{10.18653/v1/N18-1202}.
\newblock URL \url{https://www.aclweb.org/anthology/N18-1202}.

\bibitem[Raffel et~al.(2019)Raffel, Shazeer, Roberts, Lee, Narang, Matena,
  Zhou, Li, and Liu]{t5}
Colin Raffel, Noam Shazeer, Adam Roberts, Katherine Lee, Sharan Narang, Michael
  Matena, Yanqi Zhou, Wei Li, and Peter~J Liu.
\newblock Exploring the limits of transfer learning with a unified text-to-text
  transformer.
\newblock \emph{arXiv preprint arXiv:1910.10683}, 2019.

\bibitem[Rajpurkar et~al.(2016)Rajpurkar, Zhang, Lopyrev, and Liang]{squad1.1}
Pranav Rajpurkar, Jian Zhang, Konstantin Lopyrev, and Percy Liang.
\newblock Squad: 100,000+ questions for machine comprehension of text.
\newblock \emph{arXiv preprint arXiv:1606.05250}, 2016.

\bibitem[Reed(1993)]{reed1993pruning}
Russell Reed.
\newblock Pruning algorithms-a survey.
\newblock \emph{IEEE transactions on Neural Networks}, 4\penalty0 (5):\penalty0
  740--747, 1993.

\bibitem[Sajjad et~al.(2020)Sajjad, Dalvi, Durrani, and Nakov]{sajjad2020poor}
Hassan Sajjad, Fahim Dalvi, Nadir Durrani, and Preslav Nakov.
\newblock Poor man's bert: Smaller and faster transformer models.
\newblock \emph{arXiv e-prints}, pp.\  arXiv--2004, 2020.

\bibitem[Sanh et~al.(2019)Sanh, Debut, Chaumond, and Wolf]{sanh2019distilbert}
Victor Sanh, Lysandre Debut, Julien Chaumond, and Thomas Wolf.
\newblock Distilbert, a distilled version of bert: smaller, faster, cheaper and
  lighter.
\newblock \emph{arXiv preprint arXiv:1910.01108}, 2019.

\bibitem[Sanh et~al.(2020)Sanh, Wolf, and Rush]{movement-pruning}
Victor Sanh, Thomas Wolf, and Alexander Rush.
\newblock Movement pruning: Adaptive sparsity by fine-tuning.
\newblock In H.~Larochelle, M.~Ranzato, R.~Hadsell, M.~F. Balcan, and H.~Lin
  (eds.), \emph{Advances in Neural Information Processing Systems}, volume~33,
  pp.\  20378--20389. Curran Associates, Inc., 2020.
\newblock URL
  \url{https://proceedings.neurips.cc/paper/2020/file/eae15aabaa768ae4a5993a8a4f4fa6e4-Paper.pdf}.

\bibitem[Shen et~al.(2020)Shen, Dong, Ye, Ma, Yao, Gholami, Mahoney, and
  Keutzer]{shen2020q}
Sheng Shen, Zhen Dong, Jiayu Ye, Linjian Ma, Zhewei Yao, Amir Gholami,
  Michael~W Mahoney, and Kurt Keutzer.
\newblock Q-bert: Hessian based ultra low precision quantization of bert.
\newblock In \emph{Proceedings of the AAAI Conference on Artificial
  Intelligence}, volume~34, pp.\  8815--8821, 2020.

\bibitem[Simonyan \& Zisserman(2014)Simonyan and Zisserman]{vgg}
Karen Simonyan and Andrew Zisserman.
\newblock Very deep convolutional networks for large-scale image recognition.
\newblock \emph{arXiv preprint arXiv:1409.1556}, 2014.

\bibitem[Srivastava et~al.(2014)Srivastava, Hinton, Krizhevsky, Sutskever, and
  Salakhutdinov]{srivastava2014dropout}
Nitish Srivastava, Geoffrey Hinton, Alex Krizhevsky, Ilya Sutskever, and Ruslan
  Salakhutdinov.
\newblock Dropout: a simple way to prevent neural networks from overfitting.
\newblock \emph{The journal of machine learning research}, 15\penalty0
  (1):\penalty0 1929--1958, 2014.

\bibitem[Su \& Cheng(2020)Su and Cheng]{su2020sesamebert}
Ta-Chun Su and Hsiang-Chih Cheng.
\newblock Sesamebert: Attention for anywhere.
\newblock In \emph{2020 IEEE 7th International Conference on Data Science and
  Advanced Analytics (DSAA)}, pp.\  363--369. IEEE, 2020.

\bibitem[Tenney et~al.(2019)Tenney, Das, and Pavlick]{tenney2019bert}
Ian Tenney, Dipanjan Das, and Ellie Pavlick.
\newblock Bert rediscovers the classical nlp pipeline.
\newblock \emph{arXiv preprint arXiv:1905.05950}, 2019.

\bibitem[Tukan et~al.(2020)Tukan, Maalouf, Weksler, and
  Feldman]{tukan2020compressed}
Murad Tukan, Alaa Maalouf, Matan Weksler, and Dan Feldman.
\newblock Compressed deep networks: Goodbye svd, hello robust low-rank
  approximation.
\newblock \emph{arXiv preprint arXiv:2009.05647}, 2020.

\bibitem[Vaswani et~al.(2017)Vaswani, Shazeer, Parmar, Uszkoreit, Jones, Gomez,
  Kaiser, and Polosukhin]{transformer}
A~Vaswani, N~Shazeer, N~Parmar, J~Uszkoreit, L~Jones, AN~Gomez, L~Kaiser, and
  I~Polosukhin.
\newblock Attention is all you need.
\newblock In \emph{NIPS}, 2017.

\bibitem[Wang et~al.(2018)Wang, Singh, Michael, Hill, Levy, and Bowman]{glue}
Alex Wang, Amanpreet Singh, Julian Michael, Felix Hill, Omer Levy, and Samuel~R
  Bowman.
\newblock Glue: A multi-task benchmark and analysis platform for natural
  language understanding.
\newblock \emph{arXiv preprint arXiv:1804.07461}, 2018.

\bibitem[Weischedel et~al.(2011)Weischedel, Pradhan, Ramshaw, Palmer, Xue,
  Marcus, Taylor, Greenberg, Hovy, Belvin, et~al.]{ontonotes}
Ralph Weischedel, Sameer Pradhan, Lance Ramshaw, Martha Palmer, Nianwen Xue,
  Mitchell Marcus, Ann Taylor, Craig Greenberg, Eduard Hovy, Robert Belvin,
  et~al.
\newblock Ontonotes release 4.0.
\newblock \emph{LDC2011T03, Philadelphia, Penn.: Linguistic Data Consortium},
  2011.

\bibitem[Zafrir et~al.(2019)Zafrir, Boudoukh, Izsak, and
  Wasserblat]{zafrir2019q8bert}
Ofir Zafrir, Guy Boudoukh, Peter Izsak, and Moshe Wasserblat.
\newblock Q8bert: Quantized 8bit bert.
\newblock \emph{arXiv preprint arXiv:1910.06188}, 2019.

\bibitem[Zhu \& Gupta(2017)Zhu and Gupta]{zhu2017prune}
Michael Zhu and Suyog Gupta.
\newblock To prune, or not to prune: exploring the efficacy of pruning for
  model compression.
\newblock \emph{arXiv preprint arXiv:1710.01878}, 2017.

\end{thebibliography}
\bibliographystyle{iclr2022_conference}

\newpage

\appendix
\section{Appendix}

\iffalse
\subsection{Combine block implementation}\label{apx:combine_code}

    \begin{figure}[htbp]%{R}{1.00\textwidth}
        \centering
        \includegraphics[width=0.9\columnwidth]{Images/combine_code.jpg}
        \caption{Code snippet for the combine block implementation.}
        \label{fig:combine_code}
    \end{figure}
\fi

\subsection{Hyperparameter}\label{apx:hyperparameters}
\par

\subsubsection{Standalone}
    For experiments on the probing task we chose the sequence length of 128 and train the models for 3 epochs with the batch size of 32. The learning rate of 5e-5. 
    We use the same hyperparameters for SST-2, MNLI, and QQP. 
    For SQuAD dataset we use 2 gpus cards and train the model with the sequence length of 384 for 3 epochs with a batch size of 16 (8 per gpu card).
    These hyperparameters were selected considering the memory limits of the GPU. 
    
\subsubsection{Stability for fine Pruning}
    \begin{table}[htpb]
            \centering
            \begin{tabular}{|c|c|} 
                 \hline
                 Hyperparameter&
                 Value\\
                 \hline
                 \multicolumn{2}{c}{General Parameters}\\
                 \hline
                 number of gpus & 1 \\ 
                 \hline
                 per\_gpu\_train\_batch\_size & 32 \\
                 \hline
                 per\_gpu\_eval\_batch\_size & 32 \\
                 \hline
                 num\_train\_epochs & 25 \\
                 \hline
                 max\_seq\_length & 128 \\
                 \hline
                 learning\_rate & $3 \times 10 ^ {-5}$ \\
                 \hline
                 \multicolumn{2}{c}{Pruning Parameters}\\
                 \hline
                 \multirow{2}{*}{ model\_type }& masked\_bert  \\
                 &masked\_elbert\\
                 \hline
                 \multirow{2}{*}{model\_name}& masked\_bert\_hot\_25\_0.1\_trail\_k\\
                 
                 &masked\_elbert\_hot\_25\_0.1\_trail\_k\\
                 \hline
                 \multicolumn{2}{c}{Pruning Parameters}\\
                 \hline
                 warmup\_steps & 2100 \\
                 \hline
                 mask\_scores\_learning\_rate& $1 \times 10 ^ {-2}$\\
                 \hline
                 initial\_threshold& 1\\
                 \hline
                 final\_threshold &0.10\\
                 \hline
                 initial\_warmup &1\\
                 \hline
                 final\_warmup & 1\\
                 \hline
                 pruning\_method & topK\\
                 \hline
                 mask\_init & constant\\
                 \hline
                 mask\_scale & 0 \\
                 \hline
            \end{tabular}
            \caption{ 
                The hyperparameter for experiments in Section \ref{subsec:stability}. 
                For hyperparameter description refer to \url{https://github.com/huggingface/block_movement_pruning}. The hyperparameters not mentioned in the table are kept the same as there default. 
            }
            \label{tab:hyp_stability}
        \end{table}
    For our simulations on SST-2 for stability of the fine pruning stage we use the hyperparameters shown in Table \ref{tab:hyp_stability}. 

\subsubsection{Convergences of fine pruning}

    \begin{table}[htpb]
            \centering
            \begin{tabular}{|c|c|c|c|} 
                 \hline
                 Hyperparameter&
                 \multicolumn{3}{c|}{Value}\\
                 \hline
                 \multicolumn{4}{c}{General Parameters}\\
                 \hline
                 & MNLI & QQP & SQuAD\\
                 \hline
                 number of gpus & 1 & 1 & 2 \\ 
                 \hline
                 per\_gpu\_train\_batch\_size & 32 & 32 & 8 \\
                 \hline
                 per\_gpu\_eval\_batch\_size & 32 & 32 & 8 \\
                 %\hline
                 %num\_train\_epochs &  \\
                 \hline
                 max\_seq\_length & 128 & 128 & 384 \\
                 \hline
                 learning\_rate & $3 \times 10 ^ {-5}$ & $3 \times 10 ^ {-6}$ & $3 \times 10 ^ {-5}$  \\
                 \hline
                 \multicolumn{4}{c}{Pruning Parameters}\\
                 \hline
                 \multirow{2}{*}{ model\_type }& \multicolumn{3}{c|}{masked\_bert}  \\
                 &\multicolumn{3}{c|}{masked\_elbert}\\
                 \hline
                 \multirow{2}{*}{model\_name}& \multicolumn{3}{c|}{masked\_bert\_hot\_epochs\_k\_0.1}\\
                 
                 & \multicolumn{3}{c|}{masked\_elbert\_hot\_epochs\_k\_0.1}\\
                 \hline
                 \multicolumn{4}{c}{Pruning Parameters}\\
                 \hline
                 warmup\_steps & 12000& 11000&  5400 \\
                 \hline
                 mask\_scores\_learning\_rate& \multicolumn{3}{c|}{$1 \times 10 ^ {-2}$}\\
                 \hline
                 initial\_threshold& \multicolumn{3}{c|}{1}\\
                 \hline
                 final\_threshold &\multicolumn{3}{c|}{0.10}\\
                 \hline
                 initial\_warmup & 1& 2 & 1\\
                 \hline
                 final\_warmup & 1& 3 & 2\\
                 \hline
                 pruning\_method & \multicolumn{3}{c|}{topK}\\
                 \hline
                 mask\_init & \multicolumn{3}{c|}{constant}\\
                 \hline
                 mask\_scale & \multicolumn{3}{c|}{0} \\
                 \hline
            \end{tabular}
            \caption{ 
                The hyperparameter for experiments in Section \ref{subsec:convergence}. 
            }
            \label{tab:hyp_convergence}
        \end{table}
    For our simulations on convergence for MNLI, QQP and SQuAD of the fine pruning stage we use the hyperparameters shown in Table \ref{tab:hyp_convergence}.

\subsubsection{Different Pruning Approaches}
    For movement purning we adopt the hyperparameter presented in Table \ref{tab:hyp_convergence}.  The experiment on Magnitude, L\textsubscript{0} and soft movement pruning  use the hyperparameters shown in Table \ref{tab:hyp_mag}, \ref{tab:hyp_l0} and \ref{tab:hyp_mag} respectively.
    
    \begin{table}[htpb]
            \centering
            \begin{tabular}{|c|c|c|c|} 
                 \hline
                 Hyperparameter&
                 \multicolumn{3}{c|}{Value}\\
                 \hline
                 \multicolumn{4}{c}{General Parameters}\\
                 \hline
                 & MNLI & QQP & SQuAD\\
                 \hline
                 number of gpus & 1 & 1 & 2 \\ 
                 \hline
                 per\_gpu\_train\_batch\_size & 32 & 32 & 8 \\
                 \hline
                 per\_gpu\_eval\_batch\_size & 32 & 32 & 8 \\
                 \hline
                 num\_train\_epochs & 6 & 10& 10 \\
                 \hline
                 max\_seq\_length & 128 & 128 & 384 \\
                 \hline
                 learning\_rate & \multicolumn{3}{c|}{$3 \times 10 ^ {-5}$ }\\
                 \hline
                 \multicolumn{4}{c}{Pruning Parameters}\\
                 \hline
                 \multirow{2}{*}{ model\_type }& \multicolumn{3}{c|}{masked\_bert}  \\
                 &\multicolumn{3}{c|}{masked\_elbert}\\
                 \hline
                 \multirow{2}{*}{model\_name}& \multicolumn{3}{c|}{masked\_bert\_hot\_epochs\_k\_0.1}\\
                 
                 & \multicolumn{3}{c|}{masked\_elbert\_hot\_epochs\_k\_0.1}\\
                 \hline
                 \multicolumn{4}{c}{Pruning Parameters}\\
                 \hline
                 warmup\_steps & 12000& 11000&  5400 \\
                 \hline
                 mask\_scores\_learning\_rate& \multicolumn{3}{c|}{$1 \times 10 ^ {-2}$}\\
                 \hline
                 initial\_threshold& \multicolumn{3}{c|}{1}\\
                 \hline
                 final\_threshold &\multicolumn{3}{c|}{0.10}\\
                 \hline
                 initial\_warmup & 1& 2 & 1\\
                 \hline
                 final\_warmup & 1& 3 & 2\\
                 \hline
                 pruning\_method & \multicolumn{3}{c|}{Magnitude}\\
                 \hline
            \end{tabular}
            \caption{ 
                The hyperparameter for Magnitude pruning experiments in Section \ref{subsec:diff_pruning}. 
            }
            \label{tab:hyp_mag}
        \end{table}
        
    \begin{table}[htpb]
            \centering
            \begin{tabular}{|c|c|c|c|} 
                 \hline
                 Hyperparameter&
                 \multicolumn{3}{c|}{Value}\\
                 \hline
                 \multicolumn{4}{c}{General Parameters}\\
                 \hline
                 & MNLI & QQP & SQuAD\\
                 \hline
                 number of gpus & 1 & 1 & 2 \\ 
                 \hline
                 per\_gpu\_train\_batch\_size & 32 & 32 & 8 \\
                 \hline
                 per\_gpu\_eval\_batch\_size & 32 & 32 & 8 \\
                 \hline
                 num\_train\_epochs & 6& 10&  10\\
                 \hline
                 max\_seq\_length & 128 & 128 & 384 \\
                 \hline
                 learning\_rate & \multicolumn{3}{c|}{$3 \times 10 ^ {-5}$}\\
                 \hline
                 \multicolumn{4}{c}{Pruning Parameters}\\
                 \hline
                 \multirow{2}{*}{ model\_type }& \multicolumn{3}{c|}{masked\_bert}  \\
                 &\multicolumn{3}{c|}{masked\_elbert}\\
                 \hline
                 \multirow{2}{*}{model\_name}& \multicolumn{3}{c|}{masked\_bert\_hot\_epochs\_k\_0.1}\\
                 
                 & \multicolumn{3}{c|}{masked\_elbert\_hot\_epochs\_k\_0.1}\\
                 \hline
                 \multicolumn{4}{c}{Pruning Parameters}\\
                 \hline
                 warmup\_steps & 12000& 11000&  5400 \\
                 \hline
                 mask\_scores\_learning\_rate& $1 \times 10 ^ {-1}$ &$1 \times 10 ^ {-2}$ &$1 \times 10 ^ {-1}$  \\
                 \hline
                 initial\_threshold& \multicolumn{3}{c|}{1}\\
                 \hline
                 final\_threshold &\multicolumn{3}{c|}{1}\\
                 \hline
                 initial\_warmup & \multicolumn{3}{c|}{1}\\
                 \hline
                 final\_warmup & \multicolumn{3}{c|}{1}\\
                 \hline
                 pruning\_method & \multicolumn{3}{c|}{l0}\\
                 \hline
                 mask\_init & \multicolumn{3}{c|}{constant}\\
                 \hline
                 mask\_scale & \multicolumn{3}{c|}{2.197} \\
                 \hline
                 regularization & \multicolumn{3}{c|}{l0} \\
                 \hline
                 final\_lambda (BERT\textsubscript{BASE}) & 50& 50 & 175\\
                 \hline
                 final\_lambda (BERMo\textsubscript{BASE}) & 45& 50 & 175\\
                 \hline
            \end{tabular}
            \caption{ 
                The hyperparameter for L\textsubscript{0} pruning experiments in Section \ref{subsec:diff_pruning}. 
            }
            \label{tab:hyp_l0}
        \end{table}

    \begin{table}[htpb]
            \centering
            \begin{tabular}{|c|c|c|c|} 
                 \hline
                 Hyperparameter&
                 \multicolumn{3}{c|}{Value}\\
                 \hline
                 \multicolumn{4}{c}{General Parameters}\\
                 \hline
                 & MNLI & QQP & SQuAD\\
                 \hline
                 number of gpus & 1 & 1 & 2 \\ 
                 \hline
                 per\_gpu\_train\_batch\_size & 32 & 32 & 8 \\
                 \hline
                 per\_gpu\_eval\_batch\_size & 32 & 32 & 8 \\
                 \hline
                 num\_train\_epochs & 6 & 10 & 10  \\
                 \hline
                 max\_seq\_length & 128 & 128 & 384 \\
                 \hline
                 learning\_rate & $3 \times 10 ^ {-5}$ & $3 \times 10 ^ {-6}$ & $3 \times 10 ^ {-5}$  \\
                 \hline
                 \multicolumn{4}{c}{Pruning Parameters}\\
                 \hline
                 \multirow{2}{*}{ model\_type }& \multicolumn{3}{c|}{masked\_bert}  \\
                 &\multicolumn{3}{c|}{masked\_elbert}\\
                 \hline
                 \multirow{2}{*}{model\_name}& \multicolumn{3}{c|}{masked\_bert\_hot\_epochs\_k\_0.1}\\
                 
                 & \multicolumn{3}{c|}{masked\_elbert\_hot\_epochs\_k\_0.1}\\
                 \hline
                 \multicolumn{4}{c}{Pruning Parameters}\\
                 \hline
                 warmup\_steps & 12000& 11000&  5400 \\
                 \hline
                 mask\_scores\_learning\_rate& \multicolumn{3}{c|}{$1 \times 10 ^ {-2}$}\\
                 \hline
                 initial\_threshold& \multicolumn{3}{c|}{0}\\
                 \hline
                 final\_threshold &\multicolumn{3}{c|}{0.10}\\
                 \hline
                 initial\_warmup & 1& 2 & 1\\
                 \hline
                 final\_warmup & 1& 3 & 2\\
                 \hline
                 pruning\_method & \multicolumn{3}{c|}{sigmoied\_threshold}\\
                 \hline
                 mask\_init & \multicolumn{3}{c|}{constant}\\
                 \hline
                 mask\_scale & \multicolumn{3}{c|}{0} \\
                 \hline
                 regularization & \multicolumn{3}{c|}{l1} \\
                 \hline
                 final\_lambda (BERT\textsubscript{BASE}) & 200&150& 500 \\
                 \hline
                 final\_lambda (BERT\textsubscript{BASE}) & 200&150& 500 \\
                 \hline
            \end{tabular}
            \caption{ 
                The hyperparameter for soft movement pruning experiments in Section \ref{subsec:diff_pruning}. 
            }
            \label{tab:hyp_smvp}
        \end{table}

\subsubsection{Knowledge Distillation}
\begin{table}[htpb]
            \centering
            \begin{tabular}{|c|c|c|c|} 
                 \hline
                 Hyperparameter&
                 \multicolumn{3}{c|}{Value}\\
                 \hline
                 \multicolumn{4}{c}{General Parameters}\\
                 \hline
                 & MNLI & QQP & SQuAD\\
                 \hline
                 number of gpus & 1 & 1 & 2 \\ 
                 \hline
                 per\_gpu\_train\_batch\_size & 32 & 32 & 8 \\
                 \hline
                 per\_gpu\_eval\_batch\_size & 32 & 32 & 8 \\
                 %\hline
                 %num\_train\_epochs &  \\
                 \hline
                 max\_seq\_length & 128 & 128 & 384 \\
                 \hline
                 learning\_rate & $3 \times 10 ^ {-5}$ & $3 \times 10 ^ {-6}$ & $3 \times 10 ^ {-5}$  \\
                 \hline
                 \multicolumn{4}{c}{Pruning Parameters}\\
                 \hline
                 \multirow{2}{*}{ model\_type }& \multicolumn{3}{c|}{masked\_bert}  \\
                 &\multicolumn{3}{c|}{masked\_elbert}\\
                 \hline
                 \multirow{2}{*}{model\_name}& \multicolumn{3}{c|}{masked\_bert\_hot\_epochs\_k\_0.1}\\
                 
                 & \multicolumn{3}{c|}{masked\_elbert\_hot\_epochs\_k\_0.1}\\
                 \hline
                 \multicolumn{4}{c}{Pruning Parameters}\\
                 \hline
                 warmup\_steps & 12000& 11000&  5400 \\
                 \hline
                 mask\_scores\_learning\_rate& \multicolumn{3}{c|}{$1 \times 10 ^ {-2}$}\\
                 \hline
                 initial\_threshold& \multicolumn{3}{c|}{1}\\
                 \hline
                 final\_threshold &\multicolumn{3}{c|}{0.10}\\
                 \hline
                 initial\_warmup & 1& 2 & 1\\
                 \hline
                 final\_warmup & 1& 3 & 2\\
                 \hline
                 pruning\_method & \multicolumn{3}{c|}{topK}\\
                 \hline
                 mask\_init & \multicolumn{3}{c|}{constant}\\
                 \hline
                 mask\_scale & \multicolumn{3}{c|}{0} \\
                 \hline
                 \multicolumn{4}{c}{Distillation parameters}\\
                 \hline
                 teacher\_type&\multicolumn{3}{c|}{bert}\\
                 \hline
                 teacher\_name\_or\_path &  \multicolumn{3}{c|}{fine tuned bert directory for respective task} \\
                 \hline
                 alpha\_ce&\multicolumn{3}{c|}{0.1}\\
                 \hline
                 alpha\_distill&\multicolumn{3}{c|}{0.9}\\
                 \hline
            \end{tabular}
            \caption{ 
                The hyperparameter for experiments in Section \ref{subsec:knowledge_distillation}. 
            }
            \label{tab:hyp_kd}
        \end{table}
    For our simulations on Knowledge distillation on MNLI, QQP and SQuAD in the fine pruning stage we use the hyperparameters shown in Table \ref{tab:hyp_kd}.

\subsection{Different trials} \label{apx:standalone_trails}
    The results for 5 different runs for the probing task on BERT\textsubscript{BASE} and BERMo\textsubscript{BASE} can be found in Table \ref{tab:trail_standalone}.
    \begin{table}[htpb]
            \centering
        \resizebox{\columnwidth}{!}{
            \begin{tabular}{|c| c c | c c c | c c c c c|} 
                 \hline
                 \multirow{2}{*}{Model}& 
                 \multicolumn{2}{c|}{Surface}&
                 \multicolumn{3}{c|}{Syntactic}&
                 \multicolumn{5}{c|}{Semantic}\\
                 &Sentlen& 
                 WC & 
                 TreeDepth & 
                 TopConst & 
                 Bshift & 
                 Tense & 
                 Subjnum & 
                 Objnum & 
                 Somo & 
                 CoordInv\\
                 \hline
                 \multicolumn{11}{c|}{Trial 1 (Seed 63)}\\
                 \hline
                    BERT\textsubscript{BASE} &
                     $99.15$ & 
                     $84.14$&
                     $72.56$& 
                     $77.09$& 
                     $96.29$& 
                     $84.33$& 
                     $86.38$& 
                     $90.51$& 
                     $61.79$& 
                     $82.83$ \\
                    BERMo\textsubscript{BASE} &
                     $99.23$ & 
                     $83.39$ &
                     $73.14$ & 
                     $76.74$ & 
                     $96.35$ & 
                     $87.28$ & 
                     $90.55$ & 
                     $94.53$ & 
                     $64.46$ & 
                     $83.60$\\
                 \hline
                 \multicolumn{11}{c|}{Trial 2 (Seed 77)}\\
                 \hline
                    BERT\textsubscript{BASE} &
                     $99.22$ & 
                     $84.24$&
                     $72.9$& 
                     $77.38$& 
                     $96.35$& 
                     $80.78$& 
                     $86.89$& 
                     $90.51$& 
                     $61.72$& 
                     $83.25$ \\
                    BERMo\textsubscript{BASE} &
                     $99.24$ & 
                     $83.38$ &
                     $72.34$ & 
                     $76.92$ & 
                     $96.32$ & 
                     $88.17$ & 
                     $90.13$ & 
                     $93.57$ & 
                     $63.88$ & 
                     $83.15$\\
                \hline
                 \multicolumn{11}{c|}{Trial 3 (Seed 9)}\\
                 \hline
                    BERT\textsubscript{BASE} &
                     $99.23$ & 
                     Diverges&
                     $73.12$& 
                     $77.33$& 
                     $96.13$& 
                     $83.54$& 
                     $88.94$& 
                     $91.32$& 
                     $61.03$& 
                     $82.86$ \\
                    BERMo\textsubscript{BASE} &
                     $92.23$ &
                     $83.51$ & 
                     $71.21$ &
                     $76.76$ & 
                     $96.21$ & 
                     $87.07$ & 
                     $90.23$ & 
                     $92.54$ & 
                     $62.74$ & 
                     $83.47$ \\
                \hline
                 \multicolumn{11}{c|}{Trial 4 (Seed 96)}\\
                 \hline
                    BERT\textsubscript{BASE} &
                     $99.12$ & 
                     $84.11$&
                     $72.72$& 
                     $76.81$& 
                     $96.27$& 
                     $81.98$& 
                     $87.19$& 
                     $92.01$& 
                     $62.35$& 
                     $83.21$ \\
                    BERMo\textsubscript{BASE} &
                     $99.22$ & 
                     $83.55$ &
                     $72.16$ & 
                     $76.79$ & 
                     $96.27$ & 
                     $87.18$ & 
                     $90.43$ & 
                     $94.29$ & 
                     $64.69$ & 
                     $82.49$\\
                \hline
                 \multicolumn{11}{c|}{Trial 5 (Seed 39)}\\
                 \hline
                    BERT\textsubscript{BASE} &
                     $99.27$ & 
                     Diverges&
                     $72.55$& 
                     $76.72$& 
                     $96.06$& 
                     $82.24$& 
                     $87.76$& 
                     $89.21$& 
                     $60.68$& 
                     $82.60$ \\
                    BERMo\textsubscript{BASE} &
                     $99.25$ & 
                     $83.13$ &
                     $72.95$ & 
                     $77.05$ & 
                     $96.36$ & 
                     $86.40$ & 
                     $89.72$ & 
                     $93.47$ & 
                     $64.58$ & 
                     $83.10$\\
                 \hline
                 \multicolumn{11}{c|}{Mean over 5 trails}\\
                 \hline
                    BERT\textsubscript{BASE} &
                     $99.20$ & 
                     $84.16$&
                     $\textbf{72.77}$& 
                     $\textbf{77.07}$& 
                     $96.22$& 
                     $82.57$& 
                     $87.43$& 
                     $90.71$& 
                     $61.51$& 
                     $82.95$ \\
                    BERMo\textsubscript{BASE} &
                     $\textbf{99.24}$ & 
                     $83.39$ &
                     $72.36$ & 
                     $76.85$ & 
                     $\textbf{96.30}$ & 
                     $\textbf{87.22}$ & 
                     $\textbf{90.21}$ & 
                     $\textbf{93.68}$ & 
                     $\textbf{64.27}$ & 
                     $\textbf{83.16}$\\
                 \hline
                 \multicolumn{11}{c|}{Standard Deviation}\\
                 \hline
                    BERT\textsubscript{BASE} &
                     $0.0596$ & 
                     $0.0681$&
                     $0.2421$& 
                     $0.2975$& 
                     $0.1204$& 
                     $1.3874$& 
                     $0.9799$& 
                     $1.0479$& 
                     $0.6609$& 
                     $0.2750$ \\
                    BERMo\textsubscript{BASE} &
                     $0.0114$ & 
                     $0.1641$ &
                     $0.7615$ & 
                     $0.1310$ & 
                     $0.0622$ & 
                     $0.6330$ & 
                     $0.3205$ & 
                     $0.7827$ & 
                     $0.4306$ & 
                     $0.4307$\\
                 \hline
            \end{tabular}
            }
            \caption{ 
                Detailed accuracy comparison between BERT\textsubscript{BASE} and BERMo\textsubscript{BASE} on the probing task from the SentEval dataset.
            }
            \label{tab:trail_standalone}
        \end{table}

\end{document}